\documentclass{article}

\usepackage[preprint]{neurips_2026}

\usepackage[utf8]{inputenc} 
\usepackage[T1]{fontenc}    
\usepackage{hyperref}       
\usepackage{url}            
\usepackage{booktabs}       
\usepackage{amsfonts}       
\usepackage{nicefrac}       
\usepackage{microtype}      
\usepackage{xcolor}         
\usepackage{natbib}
\usepackage{subcaption}
\usepackage{graphicx}
\usepackage{url}
\title{Dendritic Neural Networks with Equilibrium Propagation}

\author{%
  Yoshimasa Kubo \\
  Department of Computer Science\\
  Lakehead University\\
  Thunder Bay, Canada\\
  \texttt{ykubo@lakeheadu.ca} \\
}

\begin{document}

\maketitle

\begin{abstract}
Equilibrium propagation (EP) is a biologically plausible alternative to backpropagation (BP), but its effectiveness can degrade in deeper and more challenging learning settings. In parallel, dendritic neural networks have demonstrated improved performance and generalization when trained with BP, suggesting that structured, biologically inspired architectures may enhance learning. In this work, we investigate the integration of dendritic neural networks with equilibrium propagation using an advanced EP framework. We evaluate the proposed dendritic EP model on MNIST, Kuzushiji-MNIST (KMNIST), and Fashion-MNIST (FMNIST), considering both shallow and deeper architectures. Our results show that dendritic EP achieves performance comparable to standard EP on simple tasks, while providing consistent improvements on more challenging datasets and deeper models. In particular, dendritic EP significantly outperforms standard EP on KMNIST and FMNIST, and approaches the performance of dendritic networks trained with backpropagation through time.To further understand these improvements, we analyze the evolution of hidden states during the free phase. We observe that dendritic EP exhibits higher activation magnitudes and more distributed hidden-state activity compared to standard EP, indicating that dendritic structure alters the internal network dynamics. These findings suggest that incorporating dendritic structure can enhance the effectiveness of biologically plausible learning algorithms, especially in regimes where standard EP struggles. Our work highlights the importance of architectural design for improving biologically inspired training methods.
\end{abstract}

\section{Introduction}
Equilibrium propagation (EP) \citep{Scellier2017,Scellier2019,Ernoult2019,Laborieux2021,Laborieux2022} is a biologically plausible training algorithm that serves as an alternative to backpropagation (BP), which is widely used for training neural networks. Recent work \citep{Laborieux2021} has shown that recurrent neural networks trained with advanced variants of EP can achieve performance competitive with those trained using BP.

Several extensions of EP have been proposed, including applications to reinforcement learning \citep{kubo2022b}, continual learning \citep{kubo2025}, models with heterogeneous time constants \citep{Kubo2026}, and architectures incorporating convolutional layers \citep{Ernoult2019,Laborieux2021}. While these studies expand the applicability of EP, they primarily focus on the learning algorithm itself rather than the design of biologically plausible neural network architectures.

In parallel, recent studies have explored dendritic neural networks trained with BP, demonstrating improved performance in tasks such as continual learning \citep{Grewal2021} and reduced overfitting \citep{Chavlis2025}. Dendritic neurons are biologically motivated; however, these approaches rely on BP, which is not biologically plausible.

In this work, we investigate the integration of dendritic neural networks with equilibrium propagation, using an advanced EP framework proposed by \citet{Laborieux2021}. We evaluate our approach on MNIST \citep{LeCun2005}, Kuzushiji-MNIST (KMNIST) \citep{Clanuwat2018}, and Fashion-MNIST (FMNIST) \citep{Xiao2017}. Our results show that the proposed dendritic EP model outperforms standard EP-based neural networks without dendritic structure and achieves performance competitive with dendritic networks trained using backpropagation through time. 

To further understand the effect of dendritic structure, we also analyze the internal dynamics of the models by visualizing hidden-state trajectories during the free phase. This analysis provides insight into how dendritic architectures influence network representations beyond performance metrics.

\section{Methods}
In this section, we will discuss equilibrium propagation, dendritic neurons, and Model and Dataset Specification.
\subsection{Equilibrium Propagation}

Equilibrium Propagation (EP) \citep{Scellier2017,Scellier2019,Ernoult2019,Laborieux2021,Laborieux2022} is a biologically plausible learning algorithm based on energy minimization. Given an input $\mathbf{x}$, the network evolves its state variables $\mathbf{s}$ toward a fixed point determined by an energy function $E(\mathbf{s};\theta)$, where $\theta$ denotes the model parameters.

In the free phase, the network evolves without any teaching signal:
\begin{equation}
\frac{d\mathbf{s}}{dt}
=
-\frac{\partial E(\mathbf{s};\theta)}{\partial \mathbf{s}}.
\end{equation}
This phase converges to a free fixed point, denoted by $\mathbf{s}^{0}$:
\begin{equation}
\mathbf{s}^{0}
=
\arg\min_{\mathbf{s}} E(\mathbf{s};\theta).
\end{equation}

In the nudged phase, the output layer is weakly driven by a loss function $\ell(\mathbf{s},\mathbf{y})$, where $\mathbf{y}$ is the target label. For a positive nudging strength $+\beta$, the dynamics are:
\begin{equation}
\frac{d\mathbf{s}}{dt}
=
-\frac{\partial E(\mathbf{s};\theta)}{\partial \mathbf{s}}
-
\beta
\frac{\partial \ell(\mathbf{s},\mathbf{y})}{\partial \mathbf{s}},
\end{equation}
which converges to a positively nudged fixed point $\mathbf{s}^{+\beta}$.

The standard two-phase EP update is then estimated as:
\begin{equation}
\Delta \theta
\propto
\frac{1}{\beta}
\left(
\frac{\partial E(\mathbf{s}^{+\beta};\theta)}{\partial \theta}
-
\frac{\partial E(\mathbf{s}^{0};\theta)}{\partial \theta}
\right).
\end{equation}

In this study, we use the symmetric nudging variant proposed by \citet{Laborieux2021}. In addition to the positive nudged phase, the network is also nudged in the opposite direction using $-\beta$:
\begin{equation}
\frac{d\mathbf{s}}{dt}
=
-\frac{\partial E(\mathbf{s};\theta)}{\partial \mathbf{s}}
+
\beta
\frac{\partial \ell(\mathbf{s},\mathbf{y})}{\partial \mathbf{s}}.
\end{equation}
This phase converges to a negatively nudged fixed point $\mathbf{s}^{-\beta}$.

The symmetric EP update is computed using a centered finite difference:
\begin{equation}
\Delta \theta
\propto
\frac{1}{2\beta}
\left(
\frac{\partial E(\mathbf{s}^{+\beta};\theta)}{\partial \theta}
-
\frac{\partial E(\mathbf{s}^{-\beta};\theta)}{\partial \theta}
\right).
\end{equation}

Compared with the standard two-phase estimator, this centered estimator reduces the bias introduced by finite nudging. This is particularly useful in deeper networks, where accurate feedback signals are important for stable credit assignment.

\subsection{Dendritic Neurons}

Biological neurons receive inputs through distinct dendritic compartments, primarily basal and apical dendrites. Basal dendrites integrate feedforward inputs from lower layers, while apical dendrites receive feedback signals from higher layers. These compartments process signals locally before integration at the soma, enabling structured and nonlinear interactions between feedforward and feedback pathways.

To model this mechanism, we introduce a dendritic neural network architecture in which each neuron receives two types of inputs: a basal (feedforward) input and an apical (feedback) input. Our implementation follows recent dendritic neural network models that represent each neuron as a collection of nonlinear dendritic branches with aggregated outputs at the soma \citep{Han2022}. In contrast to biologically detailed compartmental models, we adopt a simplified and computationally efficient formulation that is compatible with equilibrium propagation.

Formally, for a hidden layer $\mathbf{s}^{\ell}$, the basal and apical inputs are defined as:
\begin{equation}
\mathbf{b}^{\ell} = f_b\!\left( \mathbf{W}^{\ell} \mathbf{s}^{\ell-1} \right), \quad
\mathbf{a}^{\ell} = f_a\!\left( \mathbf{B}^{\ell} \mathbf{s}^{\ell+1} \right),
\end{equation}
where $\mathbf{W}^{\ell}$ and $\mathbf{B}^{\ell}$ denote the basal and apical connections, respectively, and $f_b(\cdot)$ and $f_a(\cdot)$ represent nonlinear dendritic transformations.

Each dendritic compartment consists of multiple sparse branches. Each branch connects to a subset of presynaptic neurons, applies a linear transformation followed by a nonlinearity, and produces a local response. The outputs of these branches are then aggregated to form the dendritic input. Let $z_{i,k}^{\ell}$ denote the output of the $k$-th branch associated with neuron $i$ in layer $\ell$. The basal input is computed as:
\begin{equation}
b_i^{\ell} = \frac{1}{K} \sum_{k=1}^{K} z_{i,k}^{\ell},
\end{equation}
where $K$ is the number of branches per neuron. An analogous formulation is used for the apical input.

In practice, the number of basal and apical branches, branch sparsity, and the scaling of apical feedback are treated as hyperparameters. The specific values used in our experiments are provided in Section~\ref{sec:model_spec}.

The somatic activation is obtained by combining basal and apical inputs:
\begin{equation}
\mathbf{s}^{\ell} = \sigma\!\left( \mathbf{b}^{\ell} + \alpha \mathbf{a}^{\ell} \right),
\end{equation}
where $\sigma(\cdot)$ is the activation function and $\alpha$ controls the relative strength of the apical feedback signal.

This dendritic formulation introduces structured, sparse, and nonlinear processing of both feedforward and feedback signals, while remaining computationally efficient and compatible with equilibrium propagation. Unlike prior work based on backpropagation, our approach integrates this dendritic architecture with a biologically plausible learning rule. Summary of this part is depicted in Figure \ref{fig:arch_dend_neuron}.

\begin{table}[t]
\centering
\caption{The hyper-parameters for the EP, DEP, andDBPTT are summarized in this table. Here, $\alpha 1$ refers to the learning rate for updating the weights between the input and hidden layers, $\alpha 2$ is the learning rate for updating the weights between the hidden and output layers (or another hidden layer if there are two hidden layers), and $\alpha 3$ is the learning rate for updating the weights between the hidden and output layers (if there are two hidden layers). The 'Free Phase' and 'Clamped Phase' columns specify the number of time steps used during the free and weakly clamped phases, respectively. $\beta$ is a nudging parameter for the weakly clamped phase.}
\begin{tabular}{|l|l|l|l|}
\hline
\textbf{Parameter} & \textbf{MNIST} & \textbf{KMNIST} & \textbf{FMNIST} \\ \hline

Model variants & EP / DEP / DBPTT & EP / DEP / DBPTT & EP / DEP / DBPTT \\ \hline

$\alpha_1$ & 0.04 & 0.2 / 0.5 / 0.5 & 0.3 / 0.5 / 0.5 \\ \hline
$\alpha_2$ & 0.02 & 0.08 / 0.2 / 0.2 & 0.06 / 0.2 / 0.2 \\ \hline
$\alpha_3$ & N/A & 0.02 & 0.02 \\ \hline

$\beta$ & 0.1 & 0.2 & 0.1 / 0.2 / 0.2 \\ \hline

Free phase & 60 & 120 & 120 \\ \hline
Clamped phase & 12 & 12 & 12 \\ \hline

Hidden size & 256 & 256–256 & 256–256 \\ \hline
Batch size & 64 & 64 & 64 \\ \hline

\end{tabular}
\label{tbl:hyper_param}
\end{table}

\subsection{Model and Dataset Specification}\label{sec:model_spec}

We evaluate our equilibrium propagation model with dendritic neurons (DEP) on MNIST \citep{LeCun2005}, Kuzushiji-MNIST (KMNIST) \citep{Clanuwat2018}, and Fashion-MNIST (FMNIST) \citep{Xiao2017}. We compare against two baselines: (i) a standard EP model without dendritic structure (EP), and (ii) a dendritic model trained using backpropagation through time (DBPTT).

The hyperparameters used in our experiments are summarized in Table~\ref{tbl:hyper_param}. For MNIST, we use a single hidden layer with 256 units. For KMNIST and FMNIST, we use two hidden layers (256$\times$256) for all models, reflecting the increased complexity of these datasets.

For the activation function, we use the hard sigmoid function for most configurations. However, for the EP model on FMNIST, we use the $\tanh$ activation, as we empirically observed that hard sigmoid leads to unstable training on this dataset.

For dendritic neurons, we use a fixed configuration across all datasets, consisting of 8 basal branches, 2 apical branches, a branch sparsity of 0.5, and an apical scaling factor of 0.2. These settings were chosen to balance model expressivity and computational efficiency. For nonlinear transformations within dendritic branches, we employ the rectified linear unit (ReLU) activation function.

We train all models using stochastic gradient descent (SGD) with momentum 0.9. We do not employ adaptive optimization methods such as Adam \citep{Kingma2014}, in order to maintain consistency across models and avoid introducing additional optimization-related confounding factors.

The implementation is available on github: \url{https://github.com/ykubo82/Dendritic_EP}

\section{Results}
\begin{table*}[t]
\centering
\small
\caption{Training and test accuracy (\%) for EP, dendritic EP (DEP), and dendritic BPTT (DBPTT) across datasets. Results are reported as mean $\pm$ standard deviation over multiple runs.}
\begin{tabular}{lcc|cc|cc}
\toprule
 & \multicolumn{2}{c}{EP} & \multicolumn{2}{c}{DEP} & \multicolumn{2}{c}{DBPTT} \\
\cmidrule(r){2-3} \cmidrule(r){4-5} \cmidrule(r){6-7}
Dataset & Train & Test & Train & Test & Train & Test \\
\midrule
MNIST   & 98.63 $\pm$ 0.09 & 97.45 $\pm$ 0.10 & 98.60 $\pm$ 0.03 & 97.43 $\pm$ 0.10 & 98.76 $\pm$ 0.02 & 97.87 $\pm$ 0.05 \\
KMNIST  & 96.94 $\pm$ 0.15 & 88.54 $\pm$ 0.33 & 99.30 $\pm$ 0.11 & 90.02 $\pm$ 0.27 & 99.49 $\pm$ 0.02 & 91.92 $\pm$ 0.09 \\
FMNIST  & 86.95 $\pm$ 0.23 & 85.27 $\pm$ 0.17 & 91.97 $\pm$ 0.19 & 88.52 $\pm$ 0.14 & 94.31 $\pm$ 0.04 & 89.29 $\pm$ 0.17 \\
\bottomrule
\end{tabular}

\label{tbl:results}
\end{table*}
\begin{figure}[t]
    \centering

    \begin{subfigure}{0.48\linewidth}
        \centering
        \includegraphics[width=\linewidth, height=5.5cm]{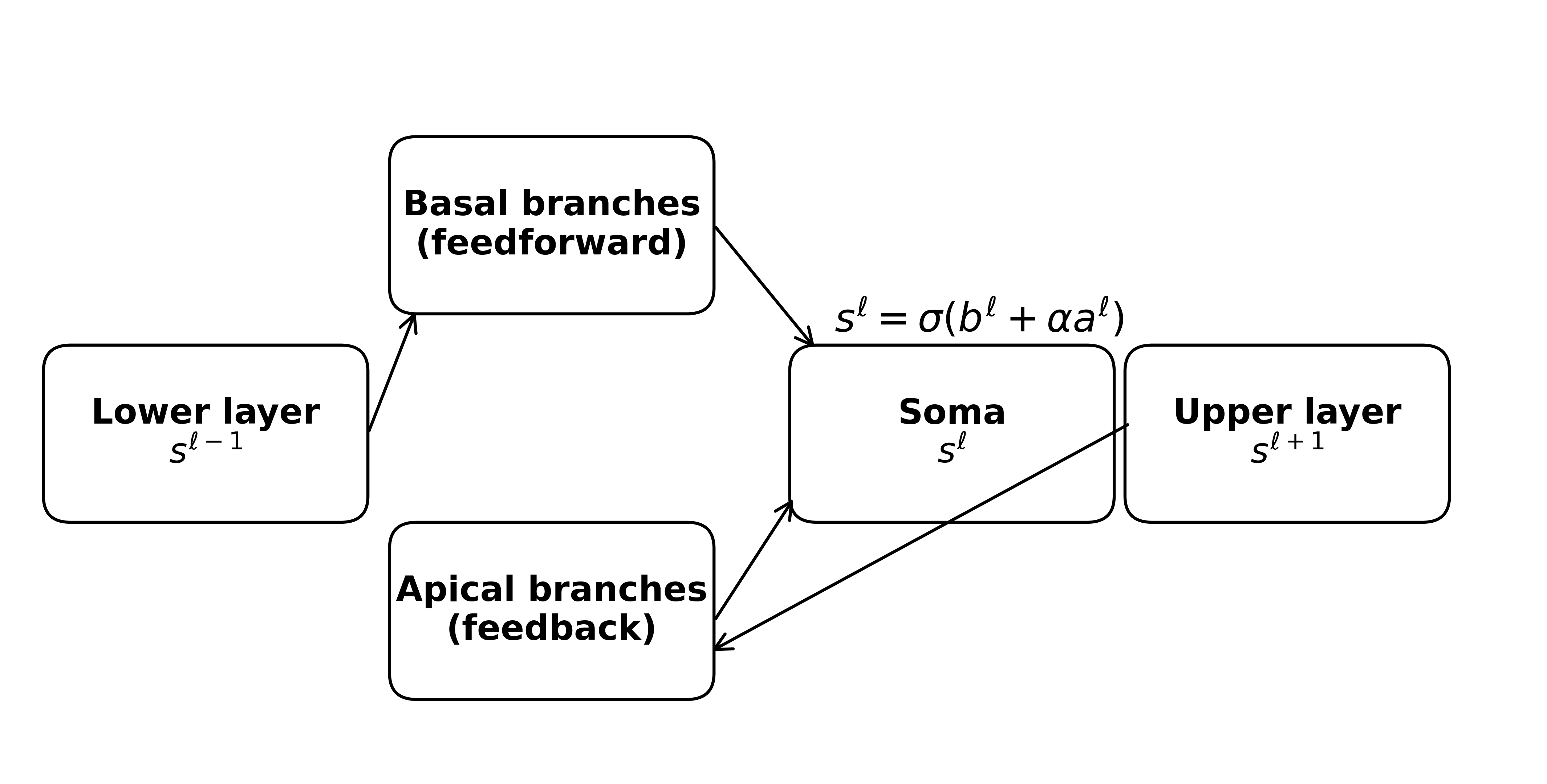}
        \caption{Dendritic neuron architecture}
        \label{fig:arch_dend_neuron}
    \end{subfigure}
    \hfill
    \begin{subfigure}{0.48\linewidth}
        \centering
        \includegraphics[width=\linewidth]{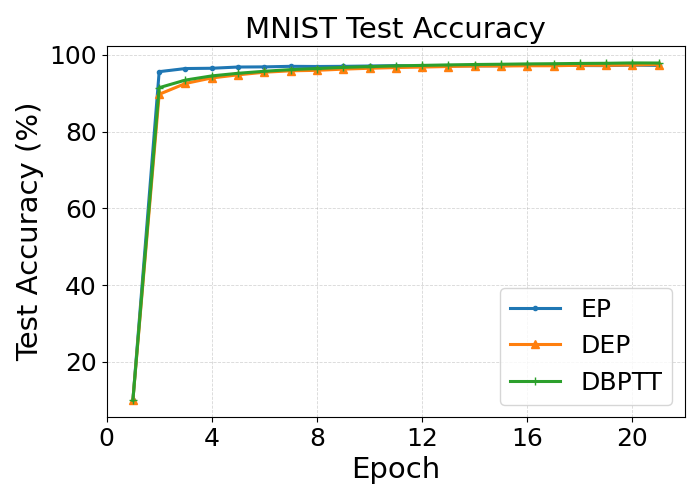}
        \caption{MNIST}
        \label{fig:MNIST_lc}
    \end{subfigure}

    \vspace{0.6em}

    \begin{subfigure}{0.48\linewidth}
        \centering
        \includegraphics[width=\linewidth]{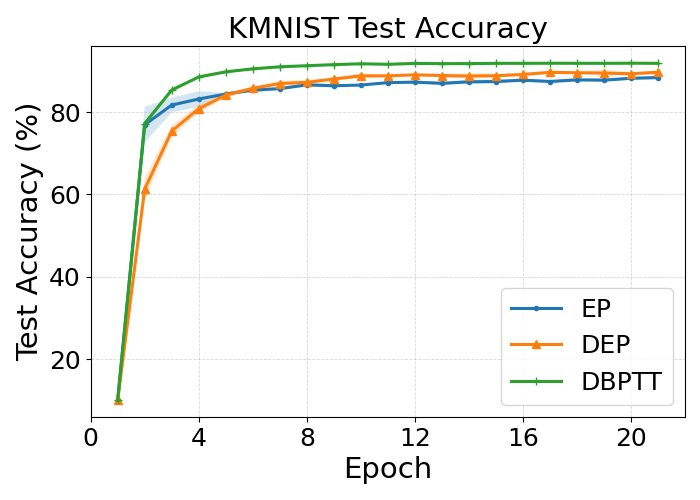}
        \caption{KMNIST}
        \label{fig:KMNIST_lc}
    \end{subfigure}
    \hfill
    \begin{subfigure}{0.48\linewidth}
        \centering
        \includegraphics[width=\linewidth]{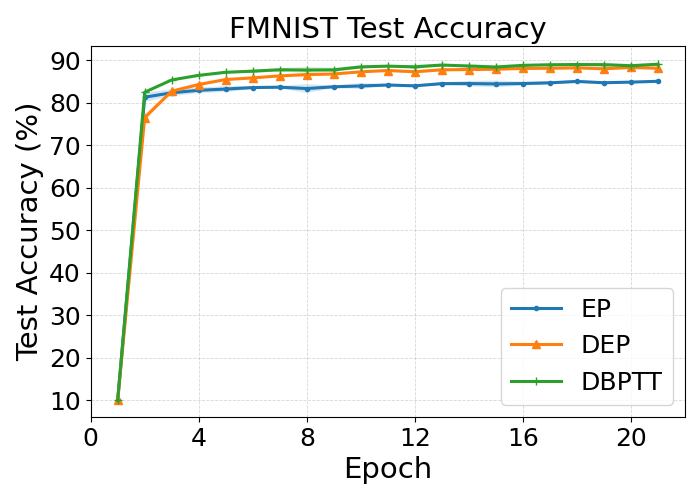}
        \caption{FMNIST}
        \label{fig:FMNIST_lc}
    \end{subfigure}

    \caption{
    \textbf{Dendritic architecture and learning dynamics.}
    (a) Illustration of the dendritic neuron, where feedforward inputs are processed through basal branches and feedback signals through apical branches before integration at the soma.
    (b–d) Test accuracy learning curves for EP, dendritic EP (DEP), and dendritic BPTT (DBPTT) on MNIST, KMNIST, and FMNIST. Shaded regions indicate standard deviation over multiple runs.
    }
    \label{fig:results}
\end{figure}
\subsection{MNIST}
The first row of Table~\ref{tbl:results} reports the performance of all models on MNIST. On this dataset, all methods achieve comparable results, with no significant differences in final test accuracy. 

Figure~\ref{fig:MNIST_lc} shows the corresponding learning curves. We observe that the standard EP model converges faster, reaching its peak performance within fewer epochs. In contrast, both dendritic models (DEP and DBPTT) require more epochs to reach their maximum accuracy, although their final performance remains similar.
\subsection{Kuzushiji-MNIST}
The second row of Table~\ref{tbl:results} summarizes the results on KMNIST. In contrast to MNIST, clearer differences between models emerge on this more challenging dataset. 

As shown in Figure~\ref{fig:KMNIST_lc}, DEP requires slightly more epochs to converge compared to the other methods. However, it achieves a substantially higher test accuracy ($90.02 \pm 0.27\%$) than standard EP ($88.54 \pm 0.33\%$), and approaches the performance of DBPTT ($91.92 \pm 0.09\%$). This suggests that incorporating dendritic structure improves performance in more complex settings while remaining competitive with backpropagation-based training.

\subsection{Fashion-MNIST}
The final row of Table~\ref{tbl:results} presents the results on FMNIST. Similar to KMNIST, we observe a clear performance gap between standard EP and the dendritic models.

Figure~\ref{fig:FMNIST_lc} shows that DEP again converges more slowly than the other models. Nevertheless, it achieves strong final performance ($88.52 \pm 0.14\%$), which is close to that of DBPTT ($89.29 \pm 0.17\%$). These results further support the effectiveness of dendritic architectures when combined with equilibrium propagation on more challenging datasets.

\subsection{States comparison}
\begin{figure}[t]
    \centering

    \begin{subfigure}{\linewidth}
        \centering
        \includegraphics[width=12cm, height=5cm]{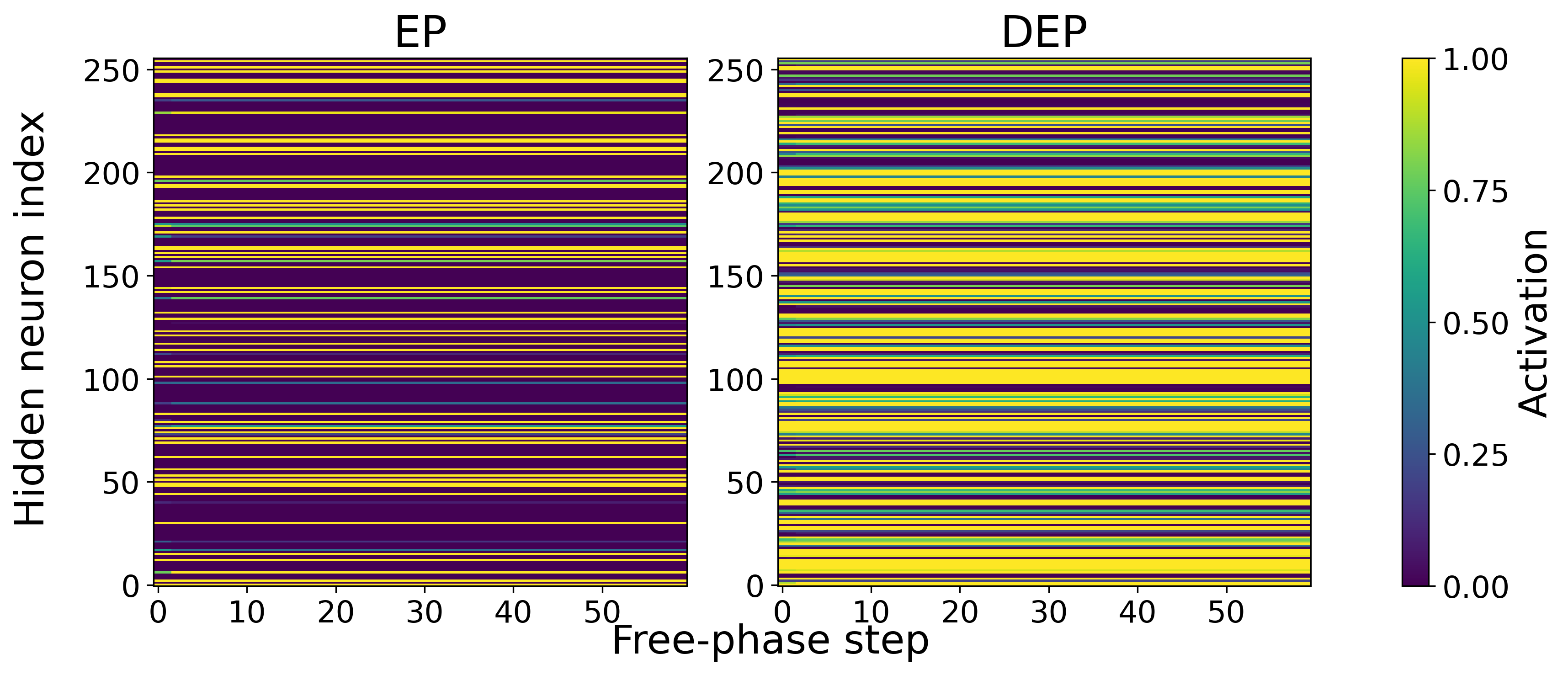}
        \caption{Representative MNIST samples with labels 7 and 2.}
        \label{fig:state_7_and_2}
    \end{subfigure}
    \vspace{0.6em}
    \begin{subfigure}{\linewidth}
        \centering
        \includegraphics[width=12cm, height=5cm]{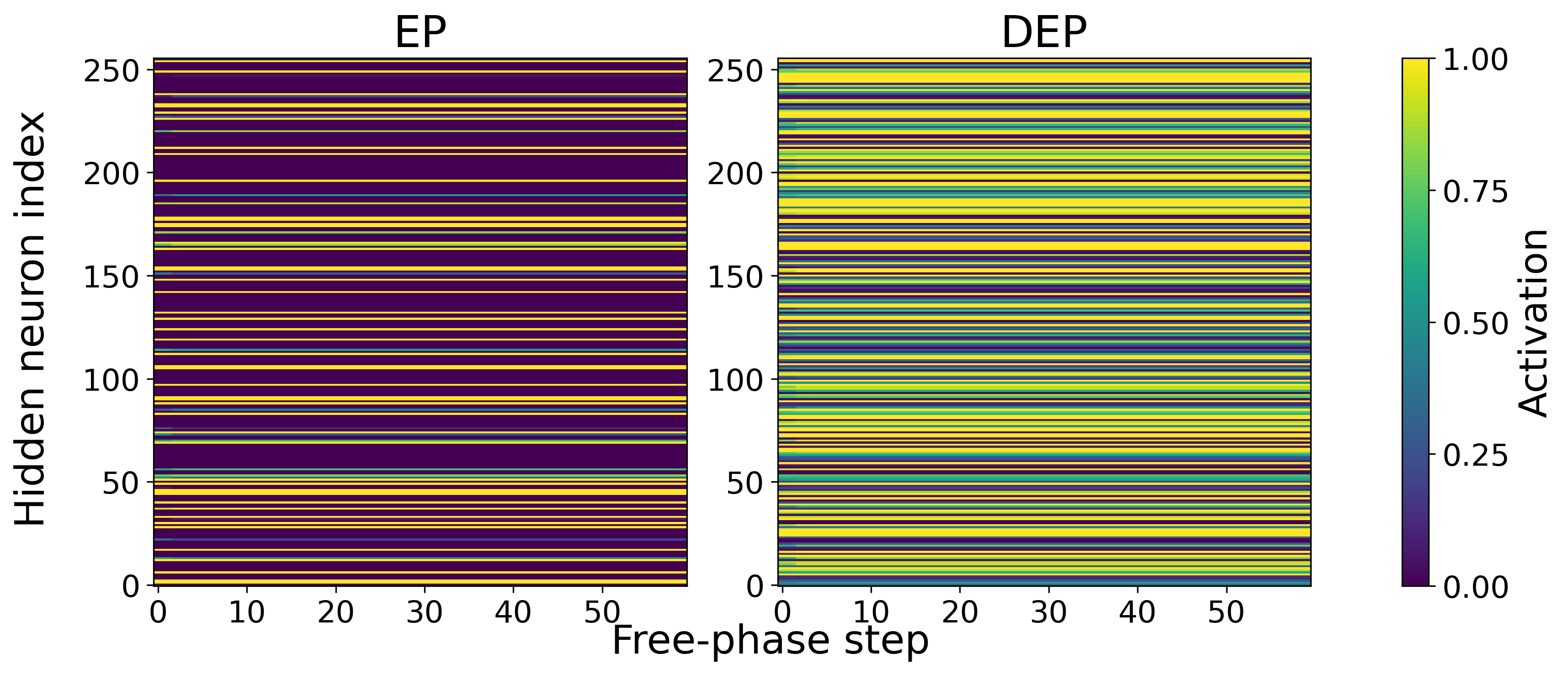}
        \caption{Representative MNIST samples with labels 1 and 0.}
        \label{fig:state_1_and_0}
    \end{subfigure}
    \caption{
    Hidden-state trajectories during the free phase for EP and DEP on representative MNIST test samples.
    DEP exhibits higher activation magnitudes and more distributed hidden-state activity compared with standard EP.
    }
    \label{fig:states_comp}
\end{figure}
To better understand the effect of dendritic neurons, we visualize the evolution of hidden states during the free phase for both EP and DEP on the MNIST dataset (Figure~\ref{fig:states_comp}). 
Across the examined examples, DEP exhibits higher activation magnitudes and engages a larger proportion of hidden neurons compared to standard EP. 
These observations suggest that incorporating dendritic structure alters the internal network dynamics, leading to more distributed hidden-state representations.

\section{Discussion}

In this work, we investigated the integration of dendritic neural network architectures with equilibrium propagation (EP). Across all datasets, the learning curves indicate that models with dendritic structure (DEP) generally converge more slowly than standard EP. This slower convergence is consistent with the increased architectural complexity introduced by dendritic branches and the additional nonlinear processing they perform.

Despite the slower learning dynamics, DEP achieves improved performance on more challenging datasets such as KMNIST and FMNIST, while remaining competitive with dendritic models trained using backpropagation through time (DBPTT). These results suggest a trade-off between convergence speed and representational capacity: incorporating dendritic structure may enhance performance in complex settings at the cost of slower optimization.

To further understand this behavior, we analyzed the evolution of hidden states during the free phase. We observed that DEP exhibits higher activation magnitudes and engages a larger proportion of hidden neurons compared to standard EP, indicating more distributed internal representations. This difference in dynamics suggests that dendritic structure alters how information is processed and propagated through the network.

One possible interpretation is that the slower convergence of DEP reflects a more gradual exploration of the energy landscape. This behavior may act as an implicit form of regularization, guiding the model toward higher-quality solutions that generalize better on more complex datasets. In particular, DEP may favor flatter regions of the energy landscape; however, validating this hypothesis requires further investigation.

Future work will explore integrating alternative biologically motivated learning rules, such as the predictive learning rule \citep{luczak2022a,luczak2022b,kubo2023}, which may provide improved learning dynamics while maintaining biological plausibility. In addition, extending this framework to spiking neural networks is a promising direction, given recent studies demonstrating the applicability of EP to spiking models \citep{Connor19a,Martin2021,Lin2024}. Combining dendritic architectures, spiking dynamics, and equilibrium-based learning may offer a more comprehensive and biologically grounded learning framework.

\subsubsection*{Acknowledgments}
This research was enabled in part by computational resources provided by the Digital Research Alliance of Canada (alliancecan.ca).

\bibliographystyle{plainnat}
\bibliography{references}

\end{document}